\documentclass[conference]{IEEEtran}

\usepackage{enumerate}
\usepackage{graphicx}
\usepackage{comment}
\usepackage{color}
\usepackage{algpseudocode}
\usepackage{amsmath}
\usepackage{wrapfig}
\usepackage{xcolor}
\usepackage{multirow}
\usepackage{tcolorbox}
\usepackage{colortbl}
\usepackage[ruled,linesnumbered]{algorithm2e}
\usepackage[left=1.62cm,right=1.62cm,top=1.9cm]{geometry}

\usepackage{url}
\usepackage{breakurl}

\begin{document}

\title{Towards Sustainable SecureML: Quantifying Carbon Footprint of Adversarial Machine Learning}

\author{\IEEEauthorblockN{Syed Mhamudul Hasan$^{1,2}$, Abdur R. Shahid$^{1,2}$, Ahmed Imteaj$^{1}$}
\IEEEauthorblockA{\textit{$^{1}$School of Computing, Southern Illinois University, Carbondale, IL, USA} \\
\textit{$^{2}$Secure and Trustworthy Intelligent Systems (SHIELD) Lab}\\
syedmhamudul.hasan@siu.edu, shahid@cs.siu.edu, imteaj@cs.siu.edu} \vspace{-1.02cm}
}

\maketitle

\begin{abstract} 

The widespread adoption of machine learning (ML) across various industries has raised sustainability concerns due to its substantial energy usage and carbon emissions. This issue becomes more pressing in adversarial ML, which focuses on enhancing model security against different network-based attacks. Implementing defenses in ML systems often necessitates additional computational resources and network security measures, exacerbating their environmental impacts. In this paper, we pioneer the first investigation into adversarial ML's carbon footprint, providing empirical evidence connecting greater model robustness to higher emissions. Addressing the critical need to quantify this trade-off, we introduce the Robustness Carbon Trade-off Index (RCTI). This novel metric, inspired by economic elasticity principles, captures the sensitivity of carbon emissions to changes in adversarial robustness. We demonstrate the RCTI through an experiment involving evasion attacks, analyzing the interplay between robustness against attacks, performance, and carbon emissions.

\end{abstract}


\IEEEpeerreviewmaketitle

\section{Introduction}\label{sec:intro}
In recent years, the rapid evolution of ML has expanded beyond the tech industry, affecting diverse sectors and heralding a new era of AI-driven innovation. However, the widespread adoption of ML raises environmental concerns due to their significant energy consumption and associated carbon emissions\cite{patterson2021carbon, wu2022sustainable}. For instance, training a single advanced language model can emit carbon equivalent to 125 round-trip flights between New York and Beijing\cite{patterson2021carbon}. The Information and Communication Technology (ICT) industry, integral to AI, is projected to account for 14\% of global emissions\cite{haldar2022environmental}, emphasizing the urgent need to address ML's environmental impact. 

\begin {figure}
    \centering
    \includegraphics[width = 0.5\textwidth]{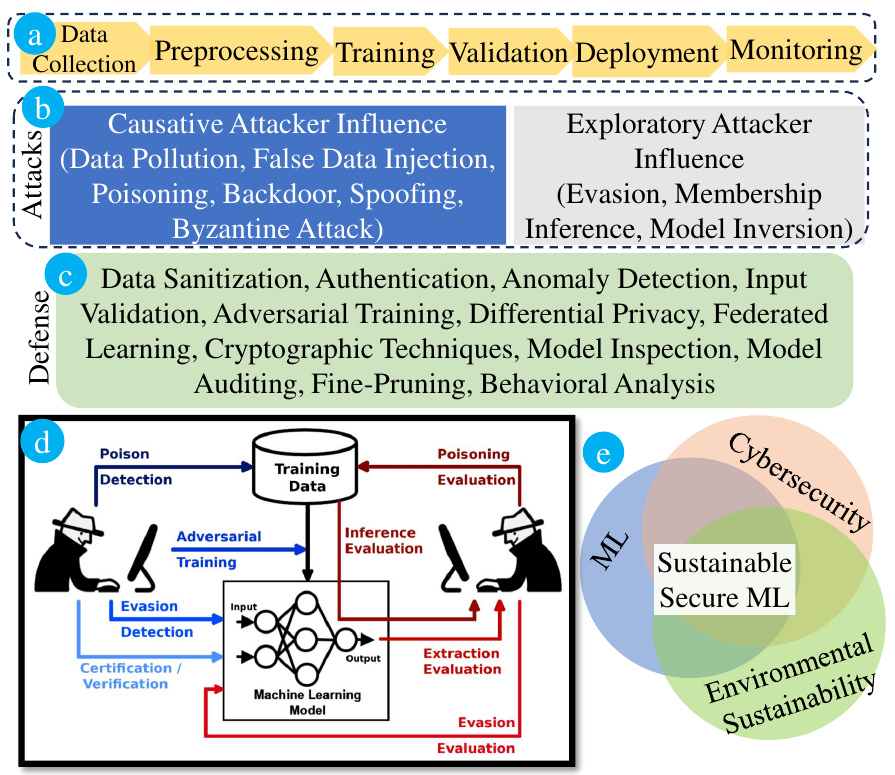}
    \vspace{-20pt}
    \caption{Sustainable SecureML: (Adversarial ML landscape) (a) An ML model lifecycle when dealing with data sources hosted on untrusted networks, (b) a classification of the adversarial attacks related to the ML lifecycle phases, (c) various defense mechanisms to defend against adversarial attacks, (d) an illustration of the attacks and their defenses, demonstrating the dynamic interplay of attack and defense tactics in untrusted network settings (figure credit: ART Toolbox\cite{art2018}), and (e) the scope of our work: the intersection of ML, cybersecurity, and environmental sustainability.}
    \label{fig:adv-ml}
    \vspace{-20pt}
\end{figure}

This research gap is particularly evident in adversarial ML — a domain at the intersection of ML and cybersecurity. Adversarial ML investigates the vulnerabilities of ML systems, crafts attack techniques for exploitation by malicious entities through trusted and untrusted networks, and develops defenses to enhance system resilience. Studies reveal how attackers can manipulate vulnerabilities in various ML phases, launching attacks to degrade model performance or inducing erroneous inferences, as illustrated in figure \ref{fig:adv-ml}. An attacker might utilize the network to execute different vulnerable or, sometimes, robust systems with zero-day vulnerabilities. As most systems need networks to communicate with each other, there might be some data alteration when data is transmitted, especially in the data collection phase. Moreover, in the data collection phase, the data can be collected from some adversarial source where the attacker intentionally sends some fake or crafted data to the training of the ML model.

To counter these attacks, several mechanisms have been explored to enhance the security of the ML systems, which often involve ML models designed to discern benign from malignant inputs. Consequently, the implementation of additional security-related computations throughout the lifecycle of adversarial ML leads to more carbon emissions. Given the direct correlation between computation and emissions, adversaries might intentionally craft attacks to escalate computational demands, further exacerbating carbon emissions. This situation underscores the urgent need to integrate environmental considerations into the development of secure ML systems, a responsibility that falls on researchers, practitioners, and policymakers alike.
\subsubsection*{Our Contributions}
This paper is the first to investigate the carbon footprint of adversarial ML systems, exploring the relationship between model robustness, environmental impact, and adversarial attacks. Central to our study is the correlation between adversarial robustness and carbon emissions, providing insights into the environmental cost of achieving a robust ML model. We use the concept of point elasticity from economics, which examines how a specific price change affects demand~\cite{mankiw2020principles}. Our contributions are as follows:
\begin{itemize}
    \item We pioneer the first work to study the carbon footprint of adversarial ML systems. Our study provides empirical evidence of a direct relationship between the robustness of an adversarial ML model and its associated carbon emissions, illustrating a crucial environmental dimension in secure and robust ML system design. 
    
    \item We propose the Robustness-Carbon Trade-off Index (RCTI), an innovative metric inspired by the well-established economic principle of point elasticity. By using a customized index, the RCTI effectively measures the relationship between a model's carbon footprint and its ability to withstand attacks from other models by suggesting the optimal improvement of resilience based on specific needs and ML model resilience.
\end{itemize}
\section{Related Work}\label{sec:related}

\subsection{Sustainable and Green ML}
Machine Learning (ML) models are highly sensitive to computational demands in terms of resources like CPUs, GPUs, and RAM, leading to increased carbon emissions. This environmental concern extends across various ML paradigms. For instance, LLMs require substantial computational power for training and testing, contributing to high carbon emissions \cite{patterson2021carbon}\cite{strubell2019energy}. Generative AI, which expands on LLM capabilities to create diverse contents, further intensifies this computational demand and associated carbon footprint \cite{Art_and_the_science_of_generative_AI}. In response, the concepts of sustainable AI and green AI have emerged to make AI computations more environmentally friendly by minimizing computational demands. This approach becomes especially pertinent as data centers significantly contribute to carbon emissions in the need for sustainable AI hardware and energy-efficient ML models \cite{MLSYS2022_462211f6}. Green AI, focusing on reduced computational costs, further contributes to sustainability, mitigating the environmental impact of ML training and testing \cite{verdecchia2023systematic}. 

However, quantifying the carbon footprint of ML systems is an ongoing challenge. Recent research efforts have aimed to assess ML systems' carbon emissions, considering factors like computational resources, data center efficiency, energy sources, model complexity, and operational efficiency \cite{wu2022sustainable, henderson2020systematic, dodge2022measuring, luccioni2023counting, kaack2022aligning}. A sustainable approach to ML, therefore, involves balancing robust model performance with minimal environmental impact. 

\subsection{Adversarial Attack on ML}

Adversarial attacks in machine learning can be categorized based on their operational phases (figure \ref{fig:adv-ml}). Adversaries utilize various sophisticated techniques to execute these attacks, including data poisoning \cite{yang2023data}, gradient-based attacks \cite{liu2023gradient}, and especially evasion attacks \cite{badr2023novel}. To counter these attacks, several mechanisms have been explored, including differential privacy \cite{ponomareva2023dp}, federated learning \cite{wei2023personalized}, cryptographic techniques \cite{zhou2023securing} and adversarial training \cite{Adversarial_Defense_Via_Local_Flatness_Regularization}. 

Evasion attacks, the center of our experiment in this paper, are especially concerning in applications where security and reliability are paramount. Two notable methods of evasion attacks are the Fast Gradient Sign Method (FGSM) and the Projected Gradient Descent (PGD). FGSM works by exploiting the gradients of the loss function with respect to the input data. It perturbs the input data in the direction of these gradients, resulting in notable changes in the model's output. FGSM's simplicity and speed make it a popular choice for demonstrating the vulnerability of neural networks to adversarial attacks. On the other hand, PGD is considered a more powerful and iterative version of FGSM. It involves applying multiple small gradient updates and projecting these updates back into the allowable input space, if necessary. Adversarial training \cite{madry2017towards} is one key method for countering evasion attacks in ML. This approach involves intentionally generating adversarial examples and including them in the training set. By doing so, the model learns to recognize and correctly classify not only the regular input data but also these adversarially perturbed inputs.

\textit{As far as our research indicates, there is no prior work that dealt with the carbon emissions of adversarial ML. This study represents the first attempt to assess the environmental implications of the robustness of an adversarial ML model.}

\section{Methodology: Measuring Robustness-
Carbon Trade-off Index of Adversarial ML}\label{sec:method}
\subsection{Design Goal}
Our objective is to develop a universal and dynamically adaptable metric for adversarial ML models, emphasizing eco-friendliness and versatility across various ML techniques. The key design goals of this metric are:

\begin{itemize}
    \item \textbf{Robustness Adjustable Adversarial ML Metric}: This metric is adjustable to the different ML systems, allowing tuning the balance between robustness in adversarial models and their carbon footprint. This feature enables quantification based on the different attack parameters and environmental factors, in our case, carbon emissions, ensuring a customized approach to sustainability in ML.

    \item \textbf{Adaptable Framework Across ML Techniques:} This framework offers an adaptable applicability to a wide spectrum of ML settings and techniques used in adversarial model training, such as Transfer Learning, Reinforcement Learning, Federated Learning, and Generative AI, facilitating the quantification and fulfillment of sustainability objectives.
\end{itemize}

\subsection{An Economic Approach to Quantify The Robustness and Carbon Emission Trade-Off}
The concept of measuring carbon emissions in adversarial machine learning (ML) presented in this paper is consistent with existing methods used for quantifying the environmental impact of ML models. We define a model's carbon emissions as a function of the system's overall energy consumption and carbon intensity, as outlined in prior research \cite{patterson2021carbon, codecarbon, gao2023review}. Let $C$ be the carbon intensity of the electricity consumed for computation, expressed in grams of $\text{CO}_2$ per kilowatt-hour (g/kWh), and $E$ be the energy consumed by the computational infrastructure to train and test a model, expressed in kilowatt-hours (kWh). Then, the total carbon emissions ($\mathcal{C}$) can be expressed in grams of $\text{CO}_2$:
\begin{equation*}\label{eq:eq_of_carbon_emission}
    \mathcal{C}  = C \times E \text{ grams} 
\end{equation*}

Moving forward, we explore the link between adversarial robustness and carbon footprint in adversarial ML models, which are designed to counter specific attacks with a attack parameter $\epsilon$, reflecting the balance between the model's resilience to attacks and its functional utility. Adversarial ML aims to optimize both model performance and robustness, a goal that introduces a complex trade-off. Operational factors in this optimization may inadvertently increase carbon emissions, thereby having an environmental impact.

A baseline model is defined as performing ML tasks without any adversarial training. Figure~\ref{fig:model_architecture} shows our baseline model architecture, where we use a convolutional neural network (CNN) to classify the MNIST dataset. Assuming the adversarial model exhibits particular performance metrics with respect to attack parameter $\epsilon$, we can define some important properties for model $\mathcal{M}$: 
\begin{itemize}
    \item $P_i( \epsilon_i)$ : Performance of the model at arbitrary $\epsilon_i$
    \item $\mathcal{C}_i( \mathcal{M}_i)$: Carbon emission of the model $\mathcal{M}_i$
    \item $\mathcal{R}_i(\epsilon_i)$: Robustness, given the attack parameter $\epsilon_i$
\end{itemize}

Considering a baseline model, $\mathcal{M}_{base}$, achieving a certain performance level, $P_{base}$, under baseline carbon emission, $\mathcal{C}_{base}$. 
Due to the vulnerability of adversarial perturbations—either introduced during training (poisoning) or testing (evasion) phases—this performance of the model $\mathcal{M}_i$ can drastically diminish to $P_i(\epsilon_i)$, with a corresponding carbon emission of $\mathcal{C}_i(\mathcal{M}_i)$. Our aim is to engineer an adversarial model $\mathcal{M}_i$ that not only enhances resilience against such attacks but also reduces environmental carbon footprints.

We quantify the robustness of $\mathcal{M}_i$ with a certain attack parameter $\epsilon_i$ as a relative improvement or degradation compared to the baseline performance as:

\vspace{-5pt}
\begin{equation}\label{eq:robustness}
    \Delta\mathcal{R}_i( \epsilon_i) = \frac{P_i( \epsilon_i) - P_{base}}{P_{base}}
    \vspace{-5pt}
\end{equation}

Where a positive $\Delta\mathcal{R}_i( \epsilon_i)$ means model $\mathcal{M}_i$ outperforms the baseline, while $\Delta\mathcal{R}_i( \epsilon_i)=0$ indicates equivalent performance, and a negative value suggests model $\mathcal{M}_i$ underperforms compared to the baseline.

\begin{table}[ht]
\caption{Elasticity of Robustness Corresponding to RCTI of an Adversarial Model}
    \label{tbl:eleasticity}
    \centering
    
    \begin{tabular}{p{1.9cm}|p{1cm}|p{4.35cm}}
        \hline
        \cellcolor{blue!30} Elasticity \break of Robustness&  \cellcolor{blue!30}$RCTI$  &   \cellcolor{blue!30}Explanation \\\hline \hline
        \cellcolor{red!60} \textbf{Eco-Critical } & \cellcolor{red!60} very high or $\infty$ & Even a minimal improvement in robustness causes a massive increase in carbon emissions. \\
        \hline 
        \cellcolor{red!20}\textbf{Eco-Costly } & \cellcolor{red!20}$> 1$ & For a given change in robustness, there is a large change in carbon emissions. \\ \hline
        
         \cellcolor{gray!15}\textbf{Eco-Neutral } &\cellcolor{gray!15} $ = 1$ &  The trade-off between robustness and carbon emissions is balanced. \\

        \hline   
        \cellcolor{green!30}\textbf{Eco-Efficient } & \cellcolor{green!30} $<1$ & Robustness can be improved with a minimal increase in carbon emissions.\\
        
        \hline 
        \cellcolor{green!95}\textbf{Eco-Ideal } & \cellcolor{green!95} $=0$ & Robustness can be improved without any increase in carbon emissions! \\
        \hline 
    \end{tabular}
\end{table}

Similarly, we can define the relative change in the carbon emission of $\mathcal{M}_i$ compared to the baseline $\mathcal{C}_{base}$ as follows:

\begin{equation}\label{eq:carbon}
    \Delta\mathcal{C}_i(\mathcal{M}_i) = \frac{\mathcal{C}_i(\mathcal{M}_i) - \mathcal{C}_{base}}{\mathcal{C}_{base}}
\end{equation}

Where, a positive $\Delta \mathcal{C}_i(\mathcal{M}_i)$ denotes a carbon emission level that exceeds the baseline. Conversely, a negative $\Delta \mathcal{C}_i(\mathcal{M}_i)$ indicates that emissions are lower than the baseline. When $\Delta \mathcal{C}_i(\mathcal{M}_i)$ is zero, it signifies that the emissions are on par with the baseline.

\begin{algorithm}[ht]
\small
\caption{Process of Measuring RCTI}\label{alg:model-rcti}
\KwData{Architectures for baseline model and $n$ adversarial robust models, attack parameter  $\{\epsilon_1, \ldots \epsilon_n\}$}
\KwResult{ List of adversarial robust models with robustness, and  carbon emissions, and RCTI values}

$\mathcal{M}\leftarrow \emptyset$: Initialize list for model data

$(M_{base}, P_{base}, \mathcal{C}_{base})\leftarrow$ Train baseline model, measure its performance (e.g., accuracy), and calculate baseline carbon emissions

\For{$i \in 1 \ldots n$}{
$(M_i, P_i( \epsilon_i), \mathcal{C}_i(\mathcal{M}_i)\leftarrow$ Train the adversarial robust $i$-the model with $\epsilon_i$, and measure its performance and emissions

$\Delta\mathcal{R}_i( \epsilon_i) \leftarrow \frac{P_i( \epsilon_i) - P_{base}}{P_{base}}$: Calculate robustness

$\Delta\mathcal{C}_i(\mathcal{M}_i) \leftarrow \frac{\mathcal{C}_i(\mathcal{M}_i) - \mathcal{C}_{base}}{\mathcal{C}_{base}}$: Measure the relative carbon emissions

$RCTI_i(\mathcal{M}_i, \epsilon_i) \leftarrow \left|\frac{\Delta \mathcal{C}_i (\mathcal{M}_i)}{\Delta\mathcal{R}_i( \epsilon_i)}\right|$: Calculate Robustness-Carbon Trade-Off Index

$\mathcal{M}\leftarrow \mathcal{M}\cup (M_i, RTCI_i(\mathcal{M}_i, \epsilon_i), \Delta\mathcal{R}_i( \epsilon_i), \Delta\mathcal{C}_i(\mathcal{M}_i) )$
}

Return $\mathcal{M}$
\end{algorithm}


By utilizing these two metrics for robustness and relative carbon emissions of an adversarial model, we quantify the relationship between adversarial robustness and carbon footprint to interpret how increasing robustness affects carbon emissions. To encapsulate this relationship, we employ a concept analogous to point elasticity—a method extensively utilized in economics to evaluate how the quantity demanded of a commodity responds to a price change at a particular point on its demand curve \cite{mankiw2020principles}. By adapting this approach to our context, we introduce the \textit{Robustness-Carbon Trade-off Index ($RCTI$)}. This index is a novel metric that quantifies the trade-off between enhancing a model's robustness and its resultant carbon emissions. For an adversarial model $\mathcal{M}_i$, the $RCTI$ is defined as the ratio of the relative change in carbon emissions to the change in model robustness. Formally, it can be expressed as follows:
\begin{equation}\label{eq:rcti}
    RCTI_i(\mathcal{M}_i,\epsilon_i)  = \left|\frac{\Delta \mathcal{C}_i(\mathcal{M}_i)}{\Delta\mathcal{R}_i( \epsilon_i)}\right|
\end{equation}

This ratio effectively captures the sensitivity of carbon emissions to alterations in model robustness. This concept of RCTI of an adversarial robust ML model $\mathcal{M}$ and attack parameter $\epsilon$ mirrors economic elasticity robustness elasticity relative to carbon emissions, as outlined in table~\ref{tbl:eleasticity}.

We define five main categories of robustness: Eco-Critical, Eco-Costly, Eco-Neutral, Eco-Efficient, and Eco-Ideal, based on how each unit of change in RCTI value impacts mentioned in Table~\ref{tbl:eleasticity}. While Eco-Ideal robustness is indeed the most desirable yet often impractical, practically, an approach would aim to find a model with eco-efficient robustness in which robustness can be improved with a minimal increase in carbon emissions. The Algorithm~\ref{alg:model-rcti} shows the process of selecting the most effective model that balances robustness with environmental sustainability and calculates robustness, carbon emissions, and RCTI for various adversarial models.

\section{Experiment and Analysis}\label{sec:experiment}

\subsection{Experimental Setup}
\subsubsection{Environment and Dataset}

\begin{wrapfigure}{r}{0.23\textwidth}
\vspace{-10pt}
  \centering
\frame{\includegraphics[width=0.23\textwidth]{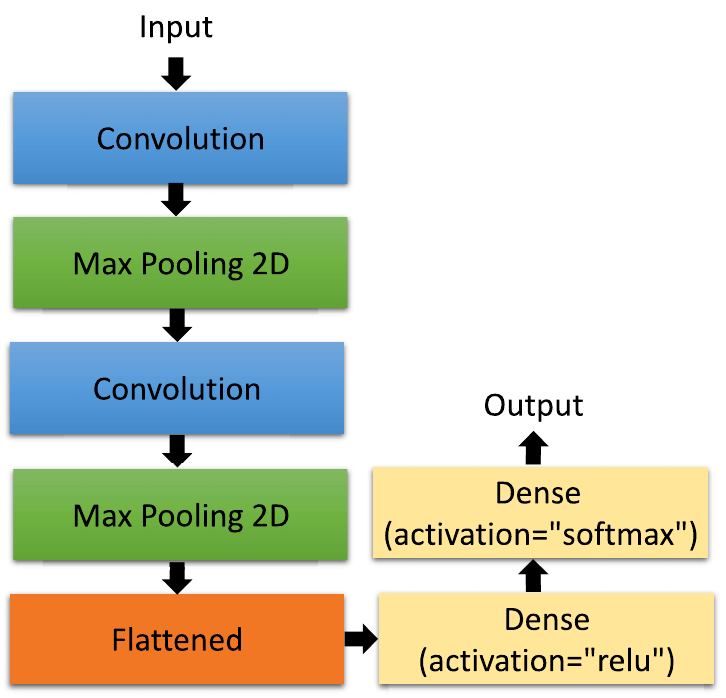}}
  \vspace{-15pt}
  \caption{{\scriptsize MNIST baseline and adversarial classifier model architecture}}
  \label{fig:model_architecture}
  \vspace{-10pt}
\end{wrapfigure}
In our experiment, we utilize Codecarbon, a Python library for quantifying carbon emissions based on metrics like CPU, RAM, GPU usage, carbon emissions, etc. \cite{codecarbon}, and the Adversarial Robustness Toolbox (ART), a widely-used Python framework for adversarial ML model implementation \cite{art2018}. We conducted the experiment on an Intel (R) Xeon (R) CPU at 2.20 GHz, 12.67834 GB of RAM, Google Colab's Linux 5.15.120, and no GPU, which is located in South Carolina, USA. To demonstrate our hypothesis on carbon emissions, we selected adversarial training with MNIST dataset.

\subsubsection{Baseline and Adversarial Model Training}

Figure~\ref{fig:model_architecture} illustrates the architecture of the baseline and adversarial models.
Initially, we we train a model without any adversarial input to establish the baseline. Subsequently, we train adversarial robust models with a varying proportion of adversarial samples. These models are different from the baseline model as they employ optimized loss functions designed for quicker convergence to local minima. We implement evasion attacks using the Fast Gradient (FG) and Projected Gradient Descent (PGD) methods, which we described in detail in the related work. In the ART implementation of FG and PGD, the hyperparameter $\epsilon$ is used to control the perturbation of the adversarial sample generation to train an adversarial robust model as well as the attack. In other words, the higher the value of $\epsilon$, the stronger the attack. We consider the $\epsilon$ range of one percent ($0.01$) to fifty percent ($0.50$) to generate different perturbations. A tracker is used to track the emissions from the various stages ($\epsilon$) of these two methods. The source code for training the baseline and robust models, including their architectures, as well as for conducting evasion attacks and adversarial training, is available on GitHub \cite{TrustedAI2023}.

\begin{table}
\small
    \centering 
    \scriptsize 
    \setlength{\tabcolsep}{.75pt} \caption{Statistics on the Hardware and Software Aspects of Baseline and Adversarial Robust Models Under Evasion Attacks}\label{tbl:stats-attacks}
    \begin{tabular}{ c|c|c|p{1cm}|p{1cm}|p{1cm}|p{1cm}|p{1cm}|p{1cm}}
\hline
\cellcolor{gray!40}\textbf{Attacks} &
\cellcolor{gray!40}\textbf{Model} & 
\cellcolor{gray!40}\textbf{$\epsilon$} & \cellcolor{gray!40}\textbf{CPU\break energy} & \cellcolor{gray!40}\textbf{Ram\break energy} & \cellcolor{gray!40}\textbf{Total\break energy} & \cellcolor{gray!40}\textbf{Accuracy (\%)} & \cellcolor{gray!40}\textbf{Duration} & \cellcolor{gray!40}\textbf{Emission}
\\
\hline

\parbox[t]{1.5mm}{\multirow{12}{*}{\rotatebox[origin=c]{90}{Fast Gradient (FG)}}} & \parbox[t]{1.5mm}{\multirow{6}{*}{\rotatebox[origin=c]{90}{Baseline}}} &  0 & 0.000032 & 3.65E-06 & 3.63E-05 & \cellcolor{cyan!98}98.42 & 2.773725 & \cellcolor{yellow!5}3.74E-06 \\
 &   &  0.1 & 0.000302 & 3.38E-05 & 3.36E-04 & \cellcolor{cyan!83}83.70 & 25.59460 & \cellcolor{yellow!20}9.59E-05 \\
  &    &   0.2 & 0.000495 & 5.54E-05 & 5.51E-04 & \cellcolor{cyan!30}30.31 & 41.97932 & \cellcolor{yellow!30}0.000157 \\
   &    &  0.3 & 0.000688 & 7.69E-05 & 7.65E-04 & \cellcolor{cyan!10}3.51 & 58.32820 & \cellcolor{yellow!40}0.000218 \\
  &     &  0.4 & 0.000885 & 9.90E-05 & 9.84E-04 & \cellcolor{cyan!4}0.46 & 75.03156 & \cellcolor{yellow!45}0.000281 \\
  &     &  0.5 & 0.001082 & 0.000121 & 0.001203 & \cellcolor{cyan!3}0.31 & 91.71211 & \cellcolor{yellow!50}0.000343 \\
\cline{2-9}

& \parbox[t]{1.5mm}{\multirow{6}{*}{\rotatebox[origin=c]{90}{Robust}}} &  
0 & 2.76E-05 & 3.08E-06 & 3.08E-06 & \cellcolor{cyan!97}97.36 & 2.340474 & \cellcolor{yellow!5}8.75E-06 \\

   &    &   0.1 & 0.000433 & 4.84E-05 & 4.84E-05 & \cellcolor{cyan!95}95.06 & 36.66551 & \cellcolor{yellow!30}0.000137  \\
  &    &   0.2 & 0.000700 & 7.83E-05 & 7.83E-05 & \cellcolor{cyan!91}91.74 & 59.35047 & \cellcolor{yellow!40}0.000222 \\
   &    &  0.3 & 0.000994 & 0.000111 & 0.000111 & \cellcolor{cyan!86}86.88 & 84.23761 & \cellcolor{yellow!45}0.000316 \\
   &    & 0.4 & 0.001304 & 0.000146 & 0.000146 & \cellcolor{cyan!40}40.16 & 110.4968 & \cellcolor{yellow!50}0.000414 \\
  &     &   0.5 & 0.001572 & 0.000176 & 0.000176 & \cellcolor{cyan!13}13.83 & 133.1732 & \cellcolor{yellow!55}0.000499  \\
\hline
\hline
\hline

\parbox[t]{1.5mm}{\multirow{12}{*}{\rotatebox[origin=c]{90}{Proj. Gradient Descent (PGD)}}} &\parbox[t]{1.5mm}{\multirow{6}{*}{\rotatebox[origin=c]{90}{Baseline}}} &  0 & 0.000108 & 1.20E-05 & 1.20E-04 & \cellcolor{cyan!97}97.88 & 2.773725 & \cellcolor{yellow!10}3.42E-05 \\
  &   &   0.1 & 0.000394 & 4.41E-05 & 4.38E-04 & \cellcolor{cyan!78}78.00 & 33.42132 & \cellcolor{yellow!30}0.000125 \\
 &   &   0.2 & 0.000594 & 6.64E-05 & 6.60E-04 & \cellcolor{cyan!7}7.00 & 50.33107 & \cellcolor{yellow!30}0.000188 \\
  &   &   0.3 & 0.000817 & 9.13E-05 & 0.000908 & \cellcolor{cyan!1}1.00 & 69.24905 & \cellcolor{yellow!40}0.000259 \\
  &   &  0.4 & 0.001032 & 0.000115 & 0.001148 & 0 & 87.48648 & \cellcolor{yellow!45}0.000328 \\
  &   &  0.5 & 0.001256 & 0.000140 & 0.001397 & 0 & 106.4692 & \cellcolor{yellow!50}0.000399  \\

\cline{2-9}

 & \parbox[t]{1.5mm}{\multirow{6}{*}{\rotatebox[origin=c]{90}{Robust}}} &  
0 & 2.76E-05 & 3.08E-06 & 3.08E-06 & \cellcolor{cyan!97}97.36 & 2.340474 & \cellcolor{yellow!5}8.75E-06 \\

 &    &  0.1 & 0.000444 & 4.96E-05 & 0.000493 & \cellcolor{cyan!98}98.00 & 37.61262 & \cellcolor{yellow!25}0.000141  \\
 &   &   0.2 & 0.000763 & 8.53E-05 & 0.000848 & \cellcolor{cyan!92}92.00 & 64.67232 & \cellcolor{yellow!30}0.000242 \\
  &   & 0.3 & 0.001061 & 0.000119 & 0.00118 & \cellcolor{cyan!87}87.00 & 89.92401 & \cellcolor{yellow!35}0.000337 \\
  &   & 0.4 & 0.001377 & 0.000154 & 0.001531 & \cellcolor{cyan!2}2.00 & 116.6996 & \cellcolor{yellow!40}0.000437 \\
  &   &   0.5 & 0.001699 & 0.000190 & 0.001889 & 0 & 143.9541 & \cellcolor{yellow!55}0.000539  \\
\hline
\multicolumn{9}{l}{Energy = kWh, Duration = Second, Emission = gCO$_2$/kWh}\\
\multicolumn{9}{l}{$^\star$The cell color intensity visually approximately reflects the values.} \\
\hline
\end{tabular}
\vspace{-15pt}
\end{table}

\subsection{Analysis of Energy, Accuracy, Time Consumption, and Emissions Under Attacks}

We provide a thorough analysis of the performance of the different machine learning models under adversarial attack scenarios in terms of energy consumption, accuracy, duration, and emissions.

\subsubsection{Energy consumption}
In both attack scenarios, the adversarial robust models consume more energy compared to the baseline models, which is expected as robust models often require more computational resources to resist attacks. For both robustly trained models, the energy increases approximately ten times with the increase of $\epsilon$. The more we add the perturbation, the more energy is needed to train the model, thus increasing the training energy requirement, as shown in the table~\ref{tbl:stats-attacks}.

\subsubsection{Accuracy}

With the increase in the $\epsilon$, the accuracy of the baseline model seems to be decreasing drastically. It holds true for robust models as well, but they are less susceptible to accuracy droppage than baseline models to a certain extent. On the table~\ref{tbl:stats-attacks}, the accuracy of the robust classifier of PGD at $\epsilon$ = 0.5 is zero. It is justifiable because when the perturbation of data is equal to or greater than the original sample, the ML model cannot function properly. That signifies a boundary in adversarial training. Focusing on the accuracy metrics, we note different impacts across different attack types. For the FG, the robust model maintains higher accuracy compared to the baseline as the attack strength increases. Similarly, in FG, the robust model has an accuracy of 95.06\% at $\epsilon = 0.1$. Furthermore, the baseline model at $\epsilon=0$ maintains a high accuracy of 98.42\%, which drops significantly as the attack strength increases. The robust model starts with a lower accuracy of 97.36\% at $\epsilon=0$ but maintains higher accuracy as the attack strength increases compared to the baseline. In both scenarios, similar behavior with an increasing level of attack parameter, $\epsilon$, correlates with increased carbon emissions from the models. This underscores a fundamental trade-off between model robustness and operational efficiency, requiring a more detailed exploration, which we presented in the subsequent analysis.

\begin{table}
\small
    \centering \scriptsize \setlength{\tabcolsep}{1pt} \caption{Robustness and Carbon Emissions Trade-Off}\label{tbl:rcti-attacks}
    \begin{tabular}{ c|p{1cm}|p{1.1cm}|p{1.1cm}|p{1.7cm}|p{2cm}}
    \hline
\cellcolor{gray!40}\textbf{Attacks} &
\cellcolor{gray!40}\textbf{$\epsilon$} & \cellcolor{gray!40}\textbf{$\Delta\mathcal{R}(\epsilon)$} & \cellcolor{gray!40}\textbf{$\Delta\mathcal{C}(\mathcal{M})$} & \cellcolor{gray!40}\textbf{$RCTI(\mathcal{M},\epsilon)$} &  \cellcolor{gray!40}\textbf{Elasticity \break of Robustness} 
\\
\hline

\parbox[t]{1.5mm}{\multirow{6}{*}{\rotatebox[origin=c]{90}{FG}}} &  0 & -0.01077 &  -1.59E-01 & 14.73085  & \cellcolor{red!20}Eco-Costly \\
 &   0.1 & 0.13570 &  4.29E-01 & 3.15769 &   \cellcolor{red!20}Eco-Costly\\
  &   0.2 & 2.02672 & 0.41401 & 0.20427 & \cellcolor{green!30}Eco-Efficient\\
   &    0.3 & 23.7521 & 0.44954 & 0.01892 & \cellcolor{green!30}Eco-Efficient \\
  &    0.4 & 86.3043 & 0.473309 & 0.005484 &  \cellcolor{green!30}Eco-Efficient\\
  &   0.5 & 43.6129 & 0.454810 & 0.010428 &  \cellcolor{green!30}Eco-Efficient\\
\hline
\parbox[t]{1.5mm}{\multirow{6}{*}{\rotatebox[origin=c]{90}{PGD}}} &  0 & -0.01077 & -1.59E-01 & 14.73085  & \cellcolor{red!20}Eco-Costly\\
  &    0.1 &  0.25641 & 0.128 & 0.49921 & \cellcolor{green!30}Eco-Efficient\\
 &   0.2 & 12.14285 & 0.28723 & 0.02365 & \cellcolor{green!30}Eco-Efficient\\
  &  0.3 & 86 & 0.301158 & 0.00350 & \cellcolor{green!30}Eco-Efficient\\
  &  0.4 & $\infty$ & 0.00043 &  $\infty$ &  \cellcolor{red!60}Eco-Critical\\
  &   0.5 & $\infty$ & 0.00053 & $\infty$ &  \cellcolor{red!60}Eco-Critical\\
  \hline
\end{tabular}
\vspace{-15pt}
\end{table}

\subsubsection{Time Consumption}
\noindent When considering the computational efficiency aspect of model performance, adversarial attacks introduce a significant computational burden. For both baseline and robust models, the duration drastically increases with the presence of adversarial samples in the dataset. For instance, in the case of the baseline model, for $\epsilon=0.1$, the computation time was 25.59460 seconds compared to its value of 2.773725 seconds for $\epsilon = 0$ (without any adversarial samples in the dataset). A similar trend is observed for the robust models. This is concerning, as adversaries might take advantage of this vulnerability to dismantle time-critical ML systems.

\subsubsection{Emission}

The total emissions, with the increase of $\epsilon$ in adversarial training, are also increasing, which is evident in the table~\ref{tbl:stats-attacks}. As it takes more computation with the increase in the perturbation parameter, we propose a correlation between robustness and emission in the adversarial machine learning domain and provide a holistic analysis of the RCTI metric in the subsequent section.

\subsection{Analysis of Robustness and Carbon Emission Trade-Off (RCTI) Under Attacks}

\begin {figure}
    \centering
    \includegraphics[width = 0.5\textwidth]{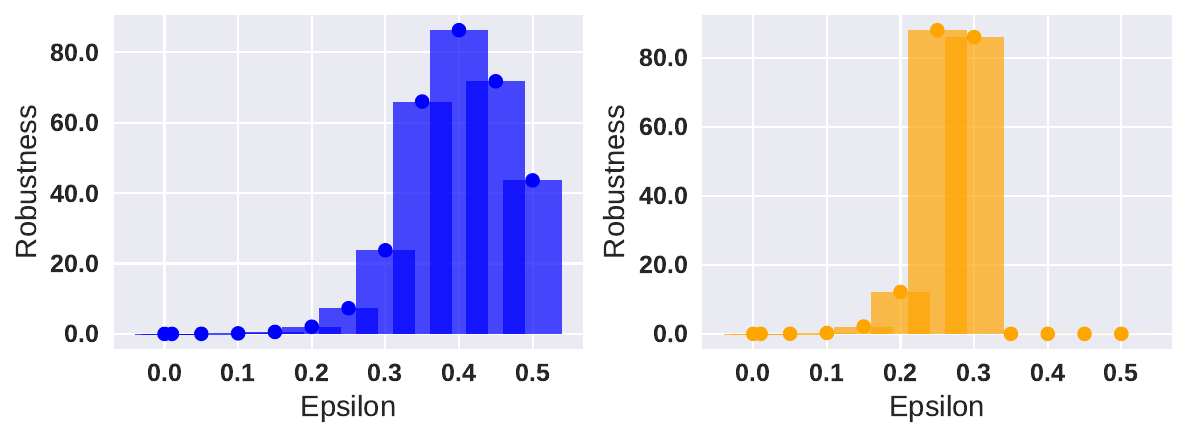}
    \vspace{-25pt}
    \caption{Robustness of adversarial models under $\epsilon$-based different evasion attacks with (left) Fast Gradient (FG) and (right) Projected Gradient Descent (PGD). For PGD, $\Delta\mathcal{R}( \epsilon)$ was set to 0 in case of $\epsilon = 0.4$ and $0.5$ for visualization purpose. Their original values were $\infty$ as shown in table \ref{tbl:rcti-attacks} }
    \label{fig:eps_vs_robustness_bar}
    \vspace{-10pt}
\end{figure}

\begin {figure}
    \centering
    \includegraphics[width = 0.5\textwidth]{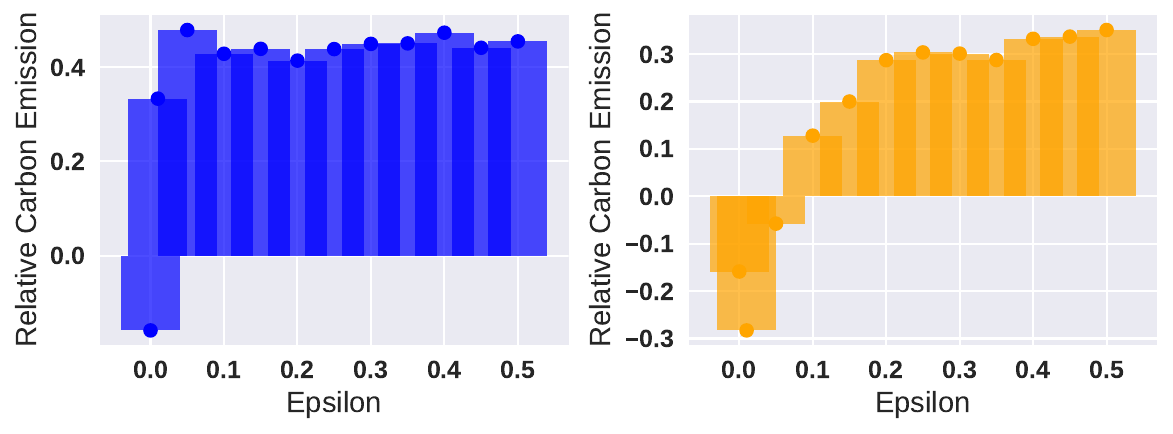}
    \vspace{-25pt}
    \caption{Carbon emission ($\Delta\mathcal{C}$) under $\epsilon$-based different evasion attacks with (left) Fast Gradient (FG) and (right) Projected Gradient Descent (PGD) }
    \label{fig:eps_vs_carbon_emission_bar}
    \vspace{-10pt}
\end{figure}

\begin {figure}
    \centering
    \includegraphics[width = 0.5\textwidth]{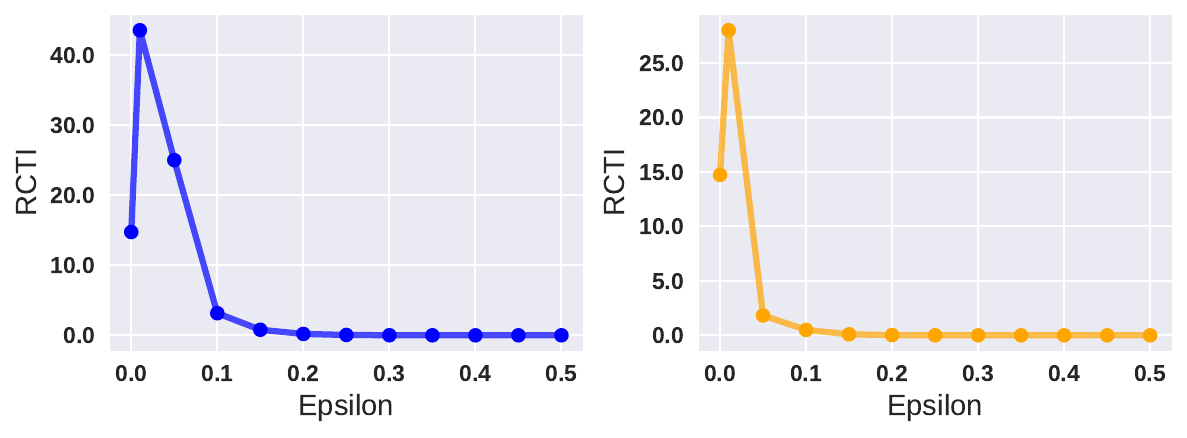}
    \vspace{-25pt}
    \caption{RCTI (Robustness-Carbon Trade-off Index) of (left) Fast Gradient (FG) and (right) Projected Gradient Descent (PGD) under different $\epsilon$. For PGD, $RCTI$ was set to 0 in case of $\epsilon = 0.4$ and $0.5$ for visualization purpose. Their original values were $\infty$ as shown in table \ref{tbl:rcti-attacks}.}
    \label{fig:eps_vs_rcti_line}
    \vspace{-10pt}
\end{figure}

Using the values extracted for both baseline and adversarial robust models, we next analyze the trade-off between robustness and adversarial attacks and the associated carbon emissions in table \ref{tbl:rcti-attacks}, and figures \ref{fig:eps_vs_robustness_bar}, \ref{fig:eps_vs_carbon_emission_bar}, and \ref{fig:eps_vs_rcti_line}. The initial insight from these results is that for both kinds of attacks, an increase in the perturbation $\epsilon$ correlates with heightened robustness $\Delta R(\epsilon)$, indicating the enhanced resilience of adversarial robust models over their baseline counterparts under evasion attacks. However, after some improvement, the robustness starts to decrease (figure \ref{fig:eps_vs_robustness_bar}). Specifically, in the case of a PGD attack, the robustness becomes $\infty$ when $\epsilon$ is 0.4 or above, coinciding with the point where the accuracy of an adversarial model regresses to zero. This is a foreseeable outcome, as with so much noisy data, a model naturally finds it increasingly difficult to generalize over the data, adversely impacting accuracy. 

We observe a similar trend in the relative change in carbon emissions $\Delta \mathcal{C}(\mathcal{M})$, as it typically rises with increasing $\epsilon$. This trend corroborates the patterns observed in table \ref{tbl:stats-attacks}. Notably, the negative $\Delta\mathcal{C}(\mathcal{M})$ values at $\epsilon$ of 0 suggest that, initially, robust models are more carbon-efficient than baseline models when assessed on the original test data.

Does this mean stronger robustness comes at a higher cost of higher carbon emissions? To answer this, let us take a look at the $RCTI$ values. Positively, the $RCTI$ value actually decreases with $\epsilon$, with lower values generally being better as they suggest a more favorable robustness-to-emission ratio. Using the elasticity of robustness metric, we can classify these different models based on whether they are eco-friendly or not. As shown in table \ref{tbl:rcti-attacks}, the elasticity of robustness shifts from eco-costly at $\epsilon = 0$ to eco-efficient as $\epsilon$ increases, reflecting a better environmental impact with increasing robustness, up to $\epsilon = 0.3$ for the PGD evasion attacks. However, at $\epsilon=0.4$, it becomes eco-critical, signaling a tipping point where the trade-off becomes environmentally unsustainable. This is encouraging news for secureML researchers, developers, and practitioners, as it underscores the potential to control carbon emissions while still advancing the development of robust adversarial ML models. Moreover, the elasticity of robustness provides an intuitive understanding of the relationship between a model's explainability, its robustness, and its environmental footprint.

\section{Conclusion}\label{sec:conclusion}

This paper presents the first attempt at assessing the carbon footprint implications in the context of adversarial machine learning. Our proposed RCTI metric can measure the connection between robustness and sustainability, which helps with figuring out how much better security for machine learning systems costs the environment and allows a customizable manner tailored to specific system demands and security settings. In our future research, we will evaluate this multi-objective metric with additional dimensions like financial costs and model fairness with the goal of sustainable robustness of ML. Expanding the evaluation across diverse datasets, model architectures, and attack types will enrich insights into this trade-off index.

\bibliographystyle{IEEEtran}
\bibliography{references}

\end{document}